\documentclass{article}
\usepackage{graphicx} % Required for inserting images
\usepackage{amsmath,amssymb,amsthm,amsfonts,bm,amsthm,color}

\def\E{{\mathbb E}}
\def\V{{\mathbb V}}
\def\X{{\mathcal X}}
\newtheorem{thm}{Theorem}

\title{Estimating the Self-Consistency of LLMs}
\author{Robert Nowak \\ University of Wisconsin-Madison}
\date{}

\begin{document}

\maketitle
Systems often repeat the same prompt to large language models (LLMs) and aggregate responses to improve reliability. Common approaches include \emph{self-consistency} or simple majority voting (sample multiple outputs and choose the mode), \emph{prompt ensembling} (rephrasing prompts to reduce wording sensitivity), and \emph{multi-agent debate} (running multiple instances and aggregating their conclusions). Such consensus methods can stabilize outputs and improve accuracy, especially on multi-step reasoning tasks \cite{miller2024adding}. This short note analyzes an estimator of the {self-consistency} of LLMs and the tradeoffs it induces under a fixed compute budget \(B=mn\), where \(m\) is the number of prompts sampled from the task distribution and \(n\) is the number of repeated LLM calls per prompt; the resulting analysis favors a rough split \(m,n\propto\sqrt{B}\). It complements recent work on self-consistency prompting that aggregates multiple sampled reasoning paths to stabilize predictions \cite{Wang2022SelfConsistency,Taubenfeld2025CISC}.

% Systems often repeat the same prompt to Large Language Models (LLMs) and aggregate the responses to improve reliability. This can be done through self-consistency or majority voting (sampling multiple outputs and selecting the most common answer), prompt ensembling (rephrasing prompts to reduce sensitivity to wording), or multi-agent debate (running multiple model instances and combining their conclusions). Such consensus methods can stabilize outputs and improve accuracy, especially in multi-step reasoning tasks  \cite{miller2024adding}.  This short note studies an estimator of the self-consistency of LLMs and the tradeoffs involved.
% This complements recent work on self-consistency prompting -- aggregating multiple sampled reasoning paths to stabilize predictions, e.g., \cite{Wang2022SelfConsistency,Taubenfeld2025CISC}.
\medskip

Consider a prompt $x$ that requires a binary response. Suppose we call an LLM $n$ times to ask if this is a positive or negative $x$.   Our analysis concerns \emph{self-consistency error}. %; i.e., take the most probable model output to be the reference answer. 
Assume the $n$ responses are independent and identically distributed and let $p(x)$ denote the probability of a positive response.   The self-consistency error of the LLM on $x$ is $${\cal E}(x)=\min\{p(x),1-p(x)\} \ .$$ 
Equivalently, taking the model’s most probable label as the reference, ${\cal E}(x)$ is the probability that an independently sampled response disagrees with that (majority) label.

The average self-consistency error over a domain of prompts $\X$ is 
$${\cal E} \ := \ \frac{1}{|\X|} \sum_{x\in \X} {\cal E}(x) \ . $$
Instead of the uniform distribution of $\X$, we can also consider a general distribution $q$ over $\X$.  In that case, the 
average self-consistency error rate over a domain of prompts $\X$ is defined to be
$${\cal E} \ := \  \sum_{x\in \X} {\cal E}(x) \, q(x) \ . $$

To estimate the self-consistency  error  for a given $x$ consider the plug-in estimator based on $\widehat p(x) = k/n$, where $k$ is the number of calls to the LLM that come back positive,
$$\widehat {\cal E}(x) \ := \ \min\{k/n,1-k/n\} \ . $$
The average error can be estimated by drawing $m$ prompts $x_1,\dots,x_m$ independently according to the distribution $q$ (possibly uniform)  from $\X$ and forming the estimator
$$\widehat {\cal E} \ := \ \frac{1}{m} \sum_{i=1}^m \widehat {\cal E}(x_i) \ . $$

\begin{thm}
    The expected squared error of the estimator $\widehat {\cal E}$ satisfies the bound
    $$\E[({\cal E}-\widehat {\cal E})^2] \ \leq \ \frac{1}{8m} + \frac{1}{\pi n} +   \frac{1}{2nm} \ . $$
\end{thm}
The first term is due to the variance incurred by the random subsample of prompts used to form the estimator.  The second term is due to the bias inherent in the estimator $\widehat {\cal E}(x)$ and the third term is due to its variance. The theorem suggests that given a budget $B$ of total calls to the LLM, the upper bound is minimized by taking $m=n=O(\sqrt{B})$ (since the third term in the bound is negligible compared to the other two terms).  The upper bound is exactly minimized by
\[
m^{*} \;=\; \sqrt{\tfrac{\pi B}{8}}, 
\qquad 
n^{*} \;=\; \sqrt{\tfrac{8B}{\pi}} \, .
\]
and these real values can be rounded to the nearest integers.
\bigskip

\begin{proof}
Assume that $n$ is even and that $p=p(x)\leq 1/2$; similar derivations can be carried out when $n$ is odd and/or $p(x)>1/2$. To bound the bias of $\widehat {\cal E}(x)$ first note that the expectation of $n \, \widehat {\cal E}(x)$ is 
\begin{eqnarray*}
    \E[\min\{k,n-k\}] & = & \sum_{k=0}^{n/2} k \binom{n}{k} p^k (1-p)^{n-k} \ + \ \sum_{n/2+1}^n (n-k) \binom{n}{k}  p^k (1-p)^{n-k} \\
    & = & \sum_{k=0}^{n} k \binom{n}{k} p^k (1-p)^{n-k} \ +  \ \sum_{n/2+1}^n (n-2k) \binom{n}{k}  p^k (1-p)^{n-k} \\
    & = & np \ + \ \sum_{n/2+1}^n (n-2k) \binom{n}{k}  p^k (1-p)^{n-k}
\end{eqnarray*}
So the bias is given by
\begin{eqnarray*}
    {\cal E}(x)-\E[\widehat {\cal E}(x)] & = & \frac{1}{n} \sum_{n/2+1}^n (2k-n) \binom{n}{k}  p^k (1-p)^{n-k} \ \geq \ 0
\end{eqnarray*}
Next we upper bound the bias.  The bound that follows is probably known, but a good reference was not found, so the derivation is included here.
First note that bias is largest when $p=1/2$ since $k>n/2$ in the sum. So we have the bound
\begin{eqnarray*}
    {\cal E}(x)-\E[\widehat {\cal E}(x)] & \leq & \frac{1}{n} \sum_{n/2+1}^n (2k-n) \binom{n}{k}  2^{-n} 
\end{eqnarray*}
The mean of a binomial distribution with parameters $n$ and $1/2$ is $n/2$, and by symmetry $\sum_{n/2+1}^n k \binom{n}{k}  2^{-n} = \sum_{k=1}{n/2} k \binom{n}{k}  2^{-n} = n/4$.  Therefore $$\frac{1}{n} \sum_{n/2+1}^n 2k \binom{n}{k}  2^{-n} = 1/2 
. $$
Next consider
\begin{eqnarray*}
    \sum_{n/2+1}^n \binom{n}{k}  2^{-n} & = & \sum_{n/2}^n \binom{n}{k}  2^{-n} - \frac{1}{2}\binom{n}{n/2}  2^{-n} \ = \ 1/2 - \frac{1}{2}\binom{n}{n/2} 2^{-n}
\end{eqnarray*}
Combining this with the previous bound, we have
\begin{eqnarray*}
    {\cal E}(x)-\E[\widehat {\cal E}(x)] & \leq & \frac{1}{2}\binom{n}{n/2} 2^{-n}
\end{eqnarray*}
This can be further  bounded using upper and lower bounds to Stirling's approximation \cite{robbins1955stirling}.  Applying these bounds we have
\begin{eqnarray*}
    \binom{n}{n/2} 2^{-n} & \leq & 2^{-n} \frac{\sqrt{2\pi n} (n/e)^n e^{\frac{1}{12n}}}{(\sqrt{2\pi n/2} ((n/2)/e)^{n/2} e^{\frac{1}{6n+1}})^2} \\
    & = & \sqrt{\frac{2}{\pi n}} e^{\frac{1}{12n}-\frac{2}{6n+1}} \ \leq \ \sqrt{\frac{2}{\pi n}}
\end{eqnarray*}
This yields the bound
\begin{eqnarray*}
    {\cal E}(x)-\E[\widehat {\cal E}(x)] & \leq &  \sqrt{\frac{1}{2\pi n}}
\end{eqnarray*}
We can bound the variance of $\widehat {\cal E}(x)$ using the fact that $\min\{p,1-p\}$ is a \mbox{$1$-Lipschitz} function in $p$.  
\begin{eqnarray*}
    \V[\widehat {\cal E}(x)] \ = \ \V[\widehat {\cal E}(x)-{\cal E}(x)] \ \leq \ \E[(\widehat {\cal E}(x)-{\cal E}(x))^2] \ \leq \ \E[(\widehat{p}(x)-p(x))^2] \ \leq \ \frac{1}{4n} \ . 
\end{eqnarray*}
Now consider estimating the overall probability of error ${\cal E}:=  \sum_{x\in \X} {\cal E}(x) \, q(x)$ using $x_1,\dots,x_m$ drawn independently according to the distribution $q$.  Define $\widehat {\cal E} := \frac{1}{m} \sum_{i=1}^m \widehat {\cal E}(x_i)$ and $\tilde {\cal E} := \frac{1}{m} \sum_{i=1}^m  {\cal E}(x_i)$.  The expected squared error of this estimator is bounded as follows.
\begin{eqnarray*}
    \E[({\cal E}-\widehat {\cal E})^2] & = & \E[({\cal E}-\tilde {\cal E} + \tilde {\cal E}- \widehat {\cal E})^2] \\ & \leq &  2\, \E[({\cal E}-\tilde {\cal E})^2] + 2 \, \E[(\tilde {\cal E}- \widehat {\cal E})^2] \\ & \leq & \frac{1}{8m} +  2\, \E[(\tilde {\cal E}- \widehat {\cal E})^2] \ ,
\end{eqnarray*}
where we use the fact that $\tilde {\cal E}$ is an unbiased estimator of ${\cal E}$ and it is the average of $m$ random variables in $[0,1/2]$; thus, its variance is at most $\frac{1}{16m}$. We next bound $\E[(\tilde {\cal E}- \widehat {\cal E})^2]$.
\begin{eqnarray*}
    \E[(\tilde {\cal E}- \widehat {\cal E})^2] & = & \frac{1}{m^2} \E\Big[\Big(\sum_{i=1}^m {\cal E}(x_i)-\widehat {\cal E}(x_i)\Big)^2\Big] \\
    & = &  \frac{1}{m^2} \E\Big[\Big(\sum_{i=1}^m {\cal E}(x_i)-\E[\widehat {\cal E}(x_i)]+ \E[\widehat {\cal E}(x_i)]-\widehat {\cal E}(x_i)\Big)^2\Big] 
\end{eqnarray*}
To simplify the notation, let $a_i = {\cal E}(x_i)-\E[\widehat {\cal E}(x_i)]$ and $b_i = \E[\widehat {\cal E}(x_i)]-\widehat {\cal E}(x_i)$, $i=1,\dots,m$.
Then
%\begin{eqnarray*}
%    \E[(\tilde e- \widehat e)^2] & = & \frac{1}{m^2} \sum_{i,j=1}^m \E[(a_i+b_i)(a_j+b_j)] \ = \ \frac{1}{m^2} \sum_{i,j=1}^m \E[a_i a_j] \ \leq \ \frac{1}{2\pi n} , 
%\end{eqnarray*}
\begin{eqnarray*}
    \E[(\tilde {\cal E}- \widehat {\cal E})^2] & = & \frac{1}{m^2} \sum_{i,j=1}^m \E[(a_i+b_i)(a_j+b_j)] \ = \ \frac{1}{m^2} \sum_{i,j=1}^m \E[a_i a_j] + \frac{1}{m^2} \sum_{i=1}^m \E[b_i^2]  \\
     & \leq & \frac{1}{2\pi n}+  \frac{1}{4nm}
\end{eqnarray*}
since $\E[a_ib_j]=\E[a_i\E[b_j|x_1,\dots,x_m]]=0$,  $a_i \leq \frac{1}{\sqrt{2\pi n}}$, and $\E[b_i^2] \leq \frac{1}{4n}$.  So we have
\begin{eqnarray*}
    \E[({\cal E}-\widehat {\cal E})^2] & \leq & \frac{1}{8m} +  \frac{1}{\pi n} +   \frac{1}{2nm} \ .
\end{eqnarray*}
\end{proof}

\paragraph{Possible Extensions.}
Beyond the mean-squared analysis of the binary self-consistency error, one can (i) derive \emph{high-probability} bounds by combining Bernstein (for $\tilde {\cal E}$) with binomial concentration for $\hat {\cal E}(x)$ and a union bound over $m$; (ii) extend to \emph{multi-class} settings by replacing ${\cal E}(x)=\min\{p,1-p\}$ with ${\cal E}(x)=1-\max_c p_c(x)$, yielding analogous $1/m$, $1/n$, and $1/(mn)$ scalings up to constants; and (iii) model correlated LLM calls via an intraclass correlation $\rho$, inflating the variance term and quantifying the erosion of majority-vote gains. These refinements preserve the optimal budget split $m^*, \, n^* \propto \sqrt{B}$ and clarify when departures from iid \ sampling (e.g., shared randomness, caching) impact the constants.

\paragraph{Acknowledgment.} This work was partially supported by NSF AI-EDGE Institute (NSF Award 2112471).

\end{document}